\def\year{2020}\relax
\documentclass[letterpaper]{article} 
\usepackage{stylecol}  
\usepackage{times}  
\usepackage{helvet} 
\usepackage{courier}  
\usepackage[hyphens]{url}  
\usepackage{graphicx} 
\usepackage{graphicx}  
\nocopyright
\usepackage{amsmath}
\usepackage{amsthm}

\def\gS{{\mathcal{S}}}
\def\gX{{\mathcal{X}}}
\def\gH{{\mathcal{H}}}
\def\gU{{\mathcal{U}}}
\def\gP{{\mathcal{P}}}
\def\gT{{\mathcal{T}}}

\def\gSe{{\mathcal{S}_{\nullsim}}}
\def\gXe{{\mathcal{X}_{\nullsim}}}
\def\vx{{\mathbf{x}}}
\def\nullsim{{\text{NULL}}}

\def\ablate{{{\textsc{Ablate}}}}

\newtheorem{theorem}{Theorem}
\newtheorem{corollary}{Corollary}
\title{Robustness Certificates for Sparse Adversarial Attacks by Randomized Ablation}
\author{Alexander Levine and Soheil Feizi\\
University of Maryland, College Park\\ \{alevine0, sfeizi\}@cs.umd.edu }
 
\begin{document}

\maketitle

\begin{abstract}
Recently, techniques have been developed to provably guarantee the robustness of a classifier to adversarial perturbations of bounded $L_1$ and $L_2$ magnitudes by using randomized smoothing: the robust classification is a consensus of base classifications on randomly noised samples where the noise is additive. In this paper, we extend this technique to the $L_0$ threat model. We propose an efficient and certifiably robust defense against sparse adversarial attacks by randomly ablating input features, rather than using additive noise. Experimentally, on MNIST, we can certify the classifications of over 50\% of images to be robust to any distortion of at most 8 pixels. This is comparable to the observed empirical robustness of unprotected classifiers on MNIST to modern $L_0$ attacks, demonstrating the tightness of the proposed robustness certificate. We also evaluate our certificate on ImageNet and CIFAR-10. Our certificates represent an improvement on those provided in a concurrent work \cite{lee2019tight} which uses random noise rather than ablation (median certificates of 8 pixels versus 4 pixels on MNIST; 16 pixels versus 1 pixel on ImageNet.) Additionally, we empirically demonstrate that our classifier is highly robust to modern sparse adversarial attacks on MNIST. Our classifications are robust, in median, to adversarial perturbations of up to 31 pixels, compared to 22 pixels reported as the state-of-the-art defense, at the cost of a slight decrease (around $2.3\%$) in the classification accuracy. Code is available at \url{https://github.com/alevine0/randomizedAblation/}.
\end{abstract}
\section{Introduction}
\begin{figure*}[ht!]
    \centering
    \includegraphics[width=.7\textwidth]{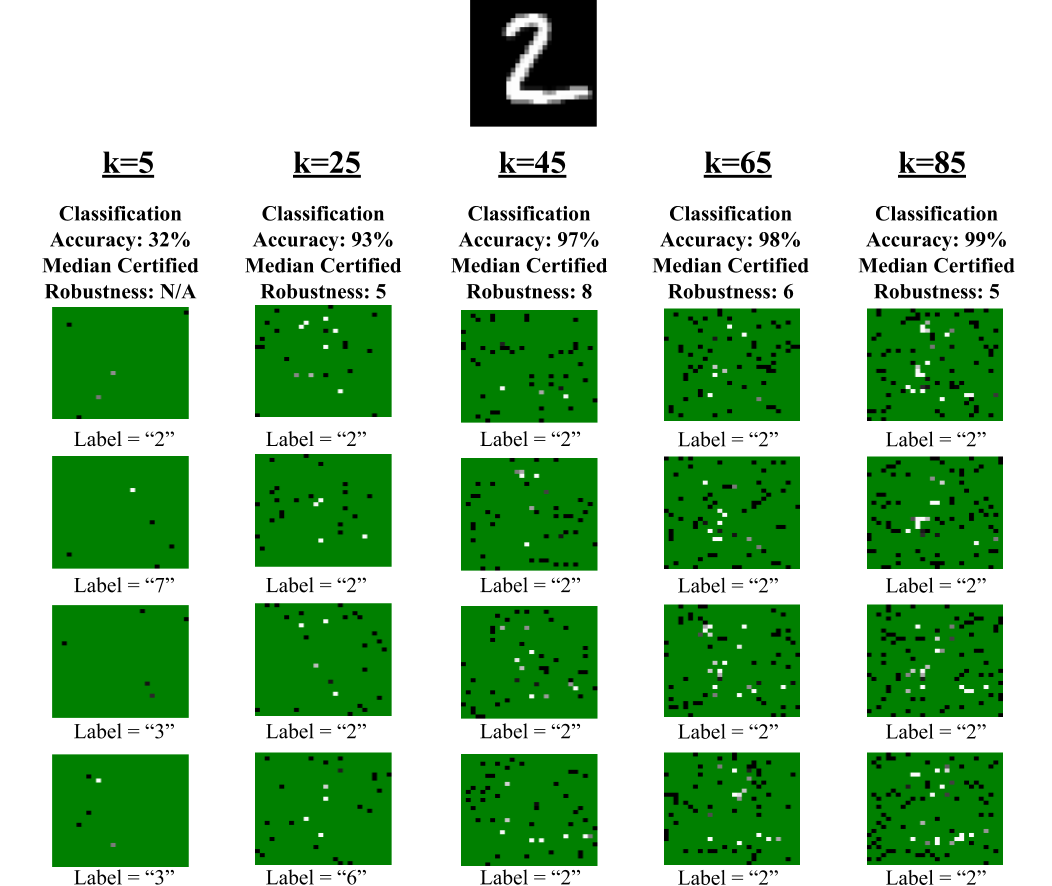}
    \caption{An illustration of our proposed certifiably robust classification scheme on MNIST. At the top, the image to be classified is shown. For randomly ablated images, we retain only $k$ out of $784$ total pixels (green pixels in these images are not used in classification). For each value of $k$, we show four randomly ablated images along with their base classifier labels. For small values of $k$, the \textit{smoothed} classifier's accuracy in the test set is low ($\sim 32\%$ for $k=5$) while the accuracy increases for moderate values of $k$ ($\sim 97\%$ for $k=45$). In each case, we compute the \textit{median certified robustness} for the smoothed classifier of the $L_0$ attack magnitude that classifications are provably protected against. The median is over the MNIST test set. For example, for $k=45$, we guarantee the robustness of our proposed method against all $L_0$ adversarial attacks that perturb 8 or fewer pixels. } 
    \label{fig:intro-fig}
\end{figure*}
Adversarial attacks, and defenses against these attacks, have been active topics of research in machine learning in recent years \cite{szegedy2013intriguing,carlini2017towards,madry2017towards}. In the case of image classification, given a classifier $f$, the goal of an adversarial attack on an image $\vx$ is to produce an image $\vx'$, such that $\vx'$ is visually similar to $\vx$, but $f$ classifies $\vx'$ differently than it classifies $\vx$. Assuming that $\vx$ is a natural image that was classified correctly, this means that the attacker can produce an image $\vx'$ which looks imperceptibly similar to this natural image, but is misclassified by $f$.\\
When designing or evaluating an adversarial attack, one must choose an objective measure of `similarity' between two images: more precisely, the goal of the attacker is to minimize $d(\vx,\vx')$, subject to $f(\vx) \neq f(\vx')$, where $d$ is a chosen distance metric. Most existing work in adversarial examples has used $L_p$ norms as distance metrics, focusing in particular on $L_\infty$ and $L_2$ norms \cite{goodfellow2014explaining,szegedy2013intriguing,madry2017towards,dong2018boosting,kurakin2018adversarial}. The $L_0$ metric, which is simply the number of pixels at which $\vx'$ differs from $\vx$, has also been the target of adversarial attacks. This metric presents a distinct challenge, because $d(\vx,\vx')$ is non-differentiable. However, both gradient-based (white-box) attacks \cite{madry2017towards,papernot2016limitations} and zeroth-order (black-box) attacks \cite{schott2019towards} have been proposed under the $L_0$ attack model. The $L_0$ attack model is the focus of this paper.\\
Several practical defenses against adversarial attacks under the $L_0$ attack model have been proposed in the last couple of years. These methods include defensive distillation \cite{papernot2016distillation}, as well as attempts to recover $\vx$ from $\vx'$ using compressed sensing \cite{bafna2018thwarting} or generative models \cite{schott2019towards,meng2017magnet}. However, as new defenses are proposed, new attacks are also developed for which these defenses are vulnerable (e.g. \cite{carlini2016defensive}). Experimental demonstrations of a defense's efficacy based on currently existing attacks do not provide a general proof of security.
In response, \textit{certifiably robust} classifiers have been developed for adversarial examples for a variety of attack models \cite{wong2018provable,gowal2018effectiveness}. For these classifiers, given an image $\vx$, it is possible to compute a radius $\rho$ such that it is guaranteed that no adversarial example $\vx'$ exists within a distance $\rho$ of $\vx$. One drawback of many of these certifiable approaches is that they can be computationally expensive since they attempt to minimize $d(\vx,\vx')$ (or its lower bound) using formal methods.\\
Recently, a relatively computationally inexpensive family of certifiably robust classifiers have been proposed which employ \textit{randomized smoothing} \cite{lecuyer2018certified,cohen2019certified,li2018second,salman2019provably}. This development has mostly been focused on the $L_1$ and $L_2$ metrics. Conceptually, these schemes work by repeatedly adding random noise to the image $\vx$, in order to create a large set of noised images. A \textit{base classifier} is then used to classify each of these noised samples, and the final robust classification is made by `majority vote.' The key insight is that, if the magnitude of the noise added to each image is much larger than the distance between $\vx$ and a potentially adversarial image $\vx'$, then any particular noised image generated from $\vx$ could have been generated from $\vx'$ with nearly equal likelihood. Then the expected number of `votes' for each class can only differ between $\vx$ and $\vx'$ by a bounded amount. Therefore, if we use a statistically sufficient number of random noise samples, and if the observed `gap' between the number of votes for the top class and the number of `votes' for any other class at $\vx$ is sufficiently large, then we can guarantee with high probability that the robust classification at $\vx'$ will be the same as it is at $\vx$. Note that the success probability can be made arbitrarily high by adding more noise samples to $\vx$ in the smoothing process.\\
In this work, we develop a certifiably robust classification scheme for the $L_0$ metric (i.e. sparse adversarial perturbations).  To guarantee the robustness of the classification against sparse adversarial attacks, we propose a novel smoothing method based on performing random \textit{ablations} on the input image, rather than adding random noise. In our proposed $L_0$ smoothing method, for each sample generated from $\vx$, a majority of pixels are randomly dropped from the image before the image is given to the base classifier. If a relatively small number $\rho$ of pixels have been adversarially corrupted (which is the case in sparse adversarial attacks), then it is highly likely that none of these pixels are present in a given ablated sample. Then, for the majority of possible random ablations, $\vx$ and $\vx'$ will give the same ablated image. Therefore, the expected number of votes for each class can only differ between $\vx$ and $\vx'$ by a bounded amount. Using this, we can prove that with high probability, the smoothed classifier will classify $\vx$ robustly against any sparse adversarial attack which is allowed to perturbed certain number of input pixels, provided that the `gap' between the number of votes for the top class and the number of `votes' for any other class at $\vx$ is sufficiently large. (See Figure \ref{fig:intro-fig})\\
Our ablation method produces significantly larger robustness guarantees compared to a more direct extension of randomized smoothing to the $L_0$ metric provided in a concurrent work by \cite{lee2019tight}: see the Discussion section for a comparison of the techniques.\\
We note that our proposed approach bears some similarities to \cite{hosseini2019dropping}, in that both works aim to defend against $L_0$ adversarial attacks by randomly ablating pixels. However, several differences exist: most notably, \cite{hosseini2019dropping} presents a \textit{practical} defense with no robustness certificate given. By contrast, the main contribution of this work is a provable guarantee of robustness to adversarial attack.\\
In summary, our contributions are as follows:
\begin{itemize}
    \item We develop a novel defense technique against sparse adversarial attacks (threat models that use the $L_0$ metric) based on randomized ablation.
    \item We characterize robustness guarantees for our proposed defense against arbitrary sparse adversarial attacks.
    \item We show the effectiveness of the proposed technique on standard datasets: MNIST, CIFAR-10, and ImageNet.
\end{itemize}
\section{Preliminaries and Notation}
We will use $\gS$ to represent the set of possible pixel values in an image. For example, in an 24-bit RGB color image, $\gS = \{0,1,...,255\}^3$, while in a binarized black-and-white image, $\gS = \{0,1\}$. We will use $\gX = \gS^d$ to represent the set of possible images, where $d$ is the number of pixels in each image. Additionally, we will use $\gSe$ to represent the set $\gS \cup \{\nullsim\}$, where $\nullsim$ is a null symbol representing the absence of information about a pixel, and $\gXe = \gSe^d$ to represent the set of images where some elements in the images may be replaced by the null symbol. Note that $\nullsim$ is {\it not} the same as a zero-valued pixel, or black. For example, if $\gS = \{0,1\}$ and $d=5$, then $[0,1,1,0,1]^T \in \gX$, while $[\nullsim,1,\nullsim,0,1]^T \in \gXe$.\\
Also, let $[d]$ represent the set of indices $\{1,...,d\}$, let $\gH(d,k) \subseteq \gP([d])$ represent all sets of $k$ unique indices in $[d]$, and let $\gU(d,k)$ represent the uniform distribution over $\gH(d,k)$. (To sample from $\gU(d,k)$ is to sample $k$ out of $d$ indices uniformly \textit{without replacement}. For example, an element sampled from $\gU(5,3)$ might be $\{2,4,5\}$.)\\
We define the operation $\ablate \in \gX \times \gH(d,k) \to \gXe$, which takes an image and a set of indices, and outputs the image, with all pixels \textit{except} those in the set replaced with the null symbol $\nullsim$. For example, $\ablate([0,1,1,0,1]^T, \{2,4,5\}) = [\nullsim,1,\nullsim,0,1]^T$\\
For images $\vx,\vx' \in \gX$, let $\|\vx-\vx'\|_0$ denote the $L_0$ distance between $\vx$ and $\vx'$, defined as the number of pixels at which $\vx$ and $\vx'$ differ. Note that we are following the convention used by \cite{carlini2017towards}, where, for a color image, the number of channels in which the images differ at a given pixel location does not matter: any difference at a pixel location (corresponding to an index in $[d]$) counts the same. This differs from \cite{papernot2016limitations}, in which channels are counted separately. Also (in a slight abuse of notation) let $\vx \ominus \vx'$ denote the set of pixel indices at which $\vx$ and $\vx'$ differ, so that $\|\vx-\vx'\|_0 = | \vx \ominus \vx'|$.\\
Finally, for multiclass classification problems, let $c$ be the number of classes.
\section{Certifiably Robust Classification Scheme}
\begin{figure}[t]
    \centering
    \includegraphics[width=0.4\textwidth]{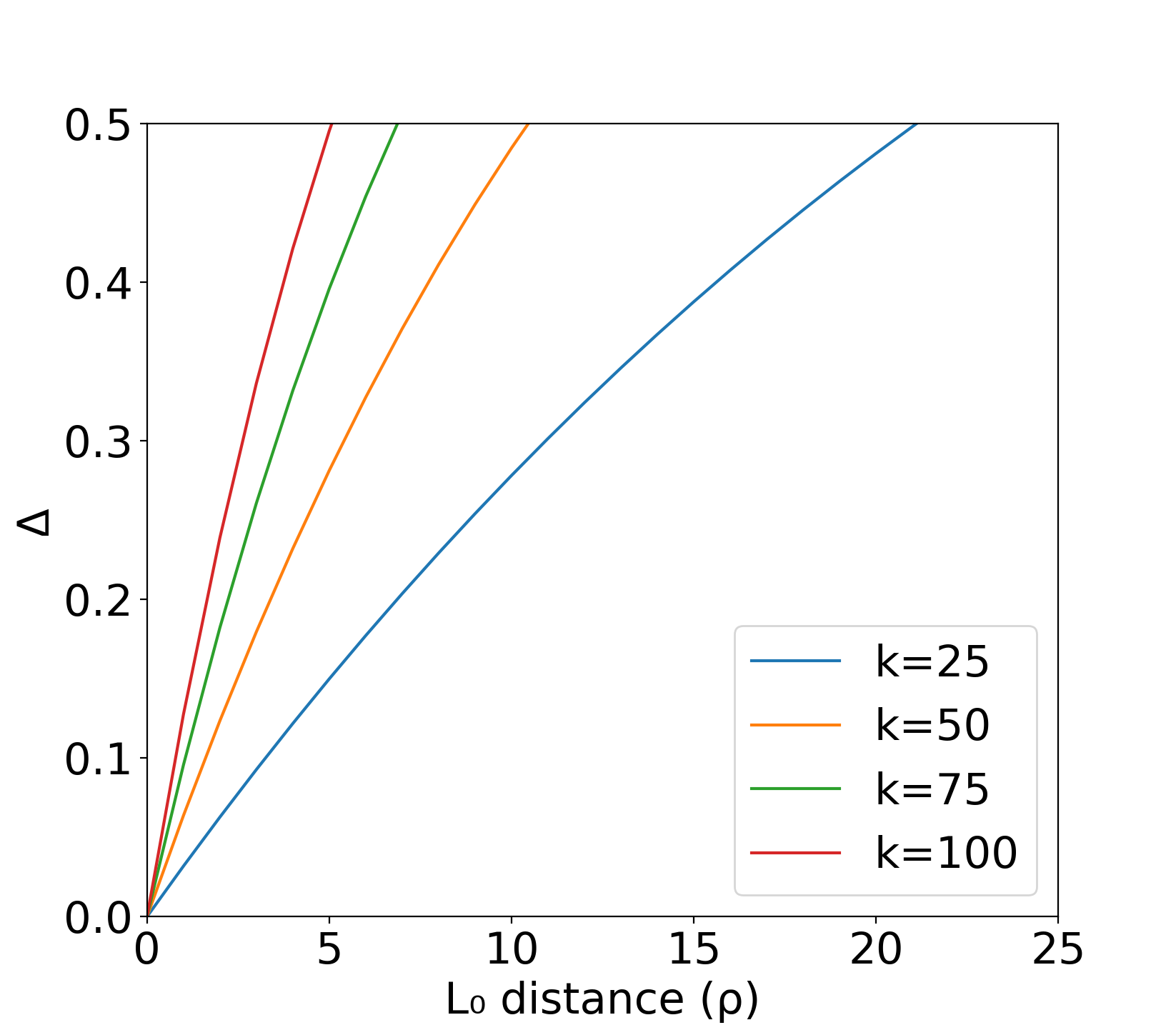}
    \caption{The bounding constant $\Delta$ from Theorem \ref{main-bound}, shown for MNIST-sized images (d=784). The constant $k$ is the number of pixels retained in each randomly ablated sample. }
    \label{fig:bound-plot}
\end{figure}

First, we note that in this section, we closely follow the notation of \cite{cohen2019certified}, using appropriate analogs between the $L_2$ smoothing scheme of that work, and the proposed $L_0$ ablation scheme of this work. In particular, let $f \in \gXe \rightarrow [c]$ denote a \textit{base classifier}, which is trained to classify images with some pixels ablated. Let $g \in \gX \rightarrow [c]$ represent a \textit{smoothed classifier}, defined as:
\begin{equation}
    g(\vx) = \arg\max_i \left[\Pr_{\gT \sim \gU(d,k)}(f(\ablate(\vx,\gT)) = i)\right]
\end{equation}
where $k$ is the \textit{retention constant}; i.e., the number of pixels retained (not ablated) from $\vx$.
In other words, $g(\vx)$ denotes the class \textit{most likely to be returned} if we first randomly ablate all but $k$ pixels from $\vx$ and then classify the resulting image with the base classifier $f$. To simplify notation, we will let $p_i(\vx)$ denote the probability that, after ablation, $f$ returns the class $i$:
\begin{equation}
    p_i(\vx) = \Pr_{\gT \sim \gU(d,k)}\left(f(\ablate(\vx,\gT)) = i \right).
\end{equation}
Thus, $g(\vx)$ can be defined simply as $ \arg\max_i \left[p_i(\vx)\right]$.\\
We first prove the following general theorem, which can be used to develop a variety of related robustness certificates.
\begin{theorem} \label{main-bound}
For images $\vx, \vx'$, with $\|\vx-\vx'\|_0 \leq \rho$, for all classes $i \in [c]$:
\begin{equation}
    \left|p_i(\vx')-p_i(\vx)\right| \leq  \Delta\\
\end{equation}    
where
\begin{equation}
\Delta = 1-\frac{\binom{d-\rho}{k}}{\binom{d}{k}}.
\end{equation}
\end{theorem}
See Figure \ref{fig:bound-plot} for a plot of how the constant $\Delta$ scales with $k$ and $\rho$.
We present a short proof of Theorem \ref{main-bound} here:
\begin{proof}
Recall that (with $\gT \sim \gU(d,k)$):
\begin{equation}
\begin{split}
p_i(\vx) &= \Pr(f(\ablate(\vx,\gT)) = i) \\
p_i(\vx') &= \Pr(f(\ablate(\vx',\gT)) = i) 
\end{split}
\end{equation}
By the law of total probability:
\begin{equation} \label{pf1totprob}
\begin{split}
&p_i(\vx)= \\
&\Pr([f(\ablate(\vx,\gT)) = i] \land [\gT \cap (\vx \ominus \vx') = \emptyset]) +\\
&\Pr([f(\ablate(\vx,\gT)) = i] \land [\gT \cap (\vx \ominus \vx') \neq \emptyset])\\
&p_i(\vx') = \\
&\Pr([f(\ablate(\vx',\gT)) = i] \land [\gT \cap (\vx \ominus \vx') = \emptyset]) +\\
&\Pr([f(\ablate(\vx',\gT)) = i] \land [\gT \cap (\vx \ominus \vx') \neq \emptyset])\\
\end{split}
\end{equation}
Note that if $\gT \cap (\vx \ominus \vx') = \emptyset$, then $\vx$ and $\vx'$ are identical at all indices in $\gT$. Then in this case, $\ablate(\vx,\gT)) = \ablate(\vx',\gT))$, which implies:
\begin{equation} \label{pf1condprob}
\begin{split}
\Pr(f(\ablate(\vx,\gT)) = i \mid \gT \cap (\vx \ominus \vx') = \emptyset)& =\\
\Pr(f(\ablate(\vx',\gT)) = i \mid \gT \cap (\vx \ominus \vx') = \emptyset) &
\end{split}
\end{equation}
Multiplying both sides of (\ref{pf1condprob}) by $\Pr(\gT \cap (\vx \ominus \vx') = \emptyset)$ gives:
\begin{equation} \label{pf1jointprob}
    \begin{split}
        &\Pr([f(\ablate(\vx,\gT)) = i] \land [\gT \cap (\vx \ominus \vx') = \emptyset]) =\\
        &\Pr([f(\ablate(\vx',\gT)) = i] \land [\gT \cap (\vx \ominus \vx') = \emptyset])
    \end{split}
\end{equation}
Substituting (\ref{pf1jointprob}) into (\ref{pf1totprob}) and rearranging yields:
\begin{equation}
    \begin{split}
        &p_i(\vx') = p_i(\vx) - \\
        & \Pr([f(\ablate(\vx,\gT)) = i] \land [\gT \cap (\vx \ominus \vx') \neq \emptyset]) +\\
         & \Pr([f(\ablate(\vx',\gT)) = i] \land [\gT \cap (\vx \ominus \vx') \neq \emptyset])\\
    \end{split}
\end{equation}
Because probabilities are non-negative, this gives:
\begin{equation}
    \begin{split}
         &p_i(\vx) - \\
         &\Pr([f(\ablate(\vx,\gT)) = i] \land [\gT \cap (\vx \ominus \vx') \neq \emptyset])\\ 
         &\leq p_i(\vx') \leq \\
       & p_i(\vx)+ \\
       &\Pr([f(\ablate(\vx',\gT)) = i] \land [\gT \cap (\vx \ominus \vx') \neq \emptyset])\\
    \end{split}
\end{equation}
By the conjunction rule, this implies:
\begin{equation} \label{pf1seclast}
    \begin{split}
        & p_i(\vx) - \Pr( \gT \cap (\vx \ominus \vx') \neq \emptyset)\\
         \leq& p_i(\vx') \leq \\
       & p_i(\vx)+ \Pr( \gT \cap (\vx \ominus \vx') \neq \emptyset)\\
    \end{split}
\end{equation}
Note that:
\begin{equation}
    \begin{split}
       \Pr( \gT \cap (\vx \ominus \vx') \neq \emptyset) =&\\ 1 - \Pr( \gT \cap (\vx \ominus \vx') = \emptyset) =& 1 - \frac{\binom{d-|\vx \ominus \vx'| }{k}}{\binom{d}{k}}
    \end{split}
\end{equation}
Where the last equality follows because $\gT$ is an uniform choice of $k$ elements from $d$: there are $\binom{d}{k}$ total ways to make this selection, $\binom{d-|\vx \ominus \vx'|}{k}$ of which contain no elements from $(\vx \ominus \vx')$.
Then:
\begin{equation} \label{pf1almlast} \begin{split}
  \Pr( \gT \cap (\vx \ominus \vx') \neq \emptyset) =  1 - \frac{\binom{d-|\vx \ominus \vx'|}{k}}{\binom{d}{k}} \\=  1 - \frac{\binom{d-\|\vx - \vx'\|_0}{k}}{\binom{d}{k}} \leq  1 - \frac{\binom{d-\rho}{k}}{\binom{d}{k}} = \Delta
     \end{split}
\end{equation}
Combining inequalities (\ref{pf1almlast}) and (\ref{pf1seclast}) gives the statement of Theorem \ref{main-bound}.
\end{proof}
\subsection{Practical Robustness Certificates}
Depending on the architecture of the base classifier, it may be infeasable to directly compute $p_i(\vx)$, and therefore to compute $g(\vx)$. However, we can instead generate a representative sample from $\gU(d,k)$, in order to bound $p_i(\vx)$ with high confidence. In particular, let $\underline{p_i(\vx)}$ represent a lower bound on $p_i(\vx)$, with $(1-\alpha)$ confidence, and let $\overline{p_i(\vx)}$ represent a similar upper bound. We first develop a certificate analogous for the $L_0$ attack to the certificate presented in \cite{cohen2019certified}: 
\begin{corollary} \label{cohenlike}
For images $\vx, \vx'$, with $\|\vx-\vx'\|_0 \leq \rho$, if:
\begin{equation}
    \underline{p_i(\vx)} - \Delta > 0.5
\end{equation}
then, with probability at least $1-\alpha$:
\begin{equation}
    g(\vx') = i
\end{equation}
\end{corollary}
\begin{proof}
With probability at least $1-\alpha$:
\begin{equation}
    .5 < \underline{p_i(\vx)} - \Delta  \leq p_i(\vx) - \Delta \leq p_i(\vx')
\end{equation}
where the final inequality is from Theorem \ref{main-bound}. Then $ g(\vx') = i$ from the definition of $g$.
\end{proof}
This bound applies directly to the true population value of $g(\vx')$, not necessarily to an empirical estimate of $g(\vx')$. Following \cite{cohen2019certified}, we therefore use a separate sampling procedure to estimate the value of the classifier $g(.)$, which itself has a bounded failure rate independent from the failure rate of the certificate, and which may abstain from classification if the top class probabilities are too similar to distinguish based on the samples. Note that by using a large number of samples, this estimation error can be made arbitrarily small. In fact, because Corollary \ref{cohenlike} is directly analogous to the condition for $L_2$ robustness presented in \cite{cohen2019certified}, we borrow both the empirical classification and the empirical certification procedures from that paper wholesale. We refer the reader to that work for details: it is sufficient to say that with these procedures, we can bound $\underline{p_i(\vx)}$ with $(1-\alpha)$ confidence and also estimate $ g(\vx')$ with $(1-\alpha)$ confidence. This is the procedure we use in our experiments.\\
Alternatively, one can instead use a certificate analogous to the certificate presented in \cite{lecuyer2018certified}. 
\begin{corollary}\label{cor2}
For images $\vx, \vx'$, with $\|\vx-\vx'\|_0 \leq \rho$, if:
\begin{equation}
    \underline{p_i(\vx)} - \Delta > \arg \max_{k\neq i} \overline{p_k(\vx)} + \Delta 
\end{equation}
then, with probability at least $1-\alpha$:
\begin{equation}
    g(\vx') = i.
\end{equation}
\end{corollary}

\begin{proof}
For each $k \neq i$:
\begin{equation}
\begin{split}
    p_k(\vx') &\leq p_k(\vx) + \Delta \leq \overline{p_k(\vx)} + \Delta \leq \arg \max_{k\neq i} \overline{p_k(\vx)} + \Delta\\ &< \underline{p_i(\vx)} - \Delta  \leq p_i(\vx) - \Delta \leq  p_i(\vx')
\end{split}
\end{equation}
where the first and last inequalities are from Theorem \ref{main-bound}.
\end{proof}
In a multi-class setting, Corollary \ref{cor2} might appear to give a tighter certificate bound. However, the upper and lower bounds on $p_j(\vx)$ must hold simultaneously for all $j$ with a total failure rate of $(1-\alpha)$. This can lead to greater estimation error if the number of classes $c$ is large. 

\subsection{Architectural and training considerations}
Similar to existing works on smoothing-based certified adversarial robustness, we train our base classifier $f$ on noisy images (i.e. ablated images), rather than training $g$ directly. For performance reasons, during training, we ablate the same pixels from all images in a minibatch. We use the same retention constant $k$ during training as at test time.
\subsubsection{Encoding $\gSe$}. We use standard CNN-based architectures for the classifier $f(.)$. However, this presents an architectural challenge: we need to be able to represent the absence of information at a pixel (the symbol $\nullsim$), as distinct from any color that can normally be encoded. Additionally, we would like the encoding of $\nullsim$ to be equally far from every possible encodable color, so that the network is not biased towards treating it as one color moreso than another. To achieve these goals, we encode images as follows: for greyscale images where pixels in $\gS$ are floating point values between zero and one (i.e. $\gS = [0,1]$), we encode $s \in \gS$ as the tuple $(s,1-s)$, and then encode $\nullsim$ as $(0,0)$. Practically, this means that we double the number of color channels from one to two, with one channel representing the original image and the other channel representing its inverse. Then, $\nullsim$ is represented as zero on both channels: this is distinct from grey $(0.5,0.5)$, white $(1,0)$, or black $(0,1)$. Notably, the values over the channels add up to one for a pixel representing any color, while it adds up to zero for a null pixel. For color images, we use the same encoding technique increasing the number of channels from 3 to 6. The resulting channels are then $(\text{red},\text{green},\text{blue},1-\text{red},1-\text{green},1-\text{blue})$, while $\nullsim$ is encoded as $(0,0,0,0,0,0)$.\footnote{On CIFAR-10, we scaled colors between 0 and 1 when using this encoding. On ImageNet, we normalized each channel to have mean 0 and standard deviation 1 before applying this encoding: in this case, the $\nullsim$ symbol is still distinct, although it is not equidistant from all other colors. }
\section{Results}
In this section, we provide experimental results of the proposed method on MNIST, CIFAR-10, and ImageNet. When reporting results, we refer to the following quantities:
\begin{itemize}
    \item The \textit{certified robustness} of a particular image $\vx$ is the maximum $\rho$ for which we can certify (with probability at least $1-\alpha$) that the smoothed classifier $g(\vx')$ will return the {\it correct} label where $\vx'$ is any adversarial perturbation of $\vx$ such that $\|\vx-\vx'\|_0 \leq \rho$. If the unperturbed classification $g(\vx)$ is itself incorrect, we define the certified robustness as N/A (Not Applicable).
    \item The \textit{certified accuracy at $\rho$} on a dataset is the fraction of images in the dataset with \textit{certified robustness} of at least $\rho$. In other words, it is the guaranteed accuracy of the classifier $g(.)$, if all images are corrupted with any $L_0$ adversarial attack of measure up to $\rho$.
    \item The \textit{median certified robustness} on a dataset is the median value of the \textit{certified robustness} across the dataset. Equivalently, it is the maximum $\rho$ for which the \textit{certified accuracy at $\rho$} is at least $0.5$. When computing this median, images which $g(.)$ misclassifies when unperturbed (i.e., \textit{certified robustness} is N/A) are counted as having $-\infty$ certified robustness. For example, if the robustness certificates of images in a dataset are \{N/A,N/A,1,2,3\}, the \textit{median certified robustness} is 1, not 2.
    \item The \textit{classification accuracy} on a dataset is the fraction of images on which our empirical estimation of $g(.)$ returns the correct class label, and does not abstain. 
    \item The \textit{empirical adversarial attack magnitude} of a particular image $\vx$ is the minimum $\rho$ for which an adversarial attack can find an adversarial example $\vx'$ such that $\|\vx-\vx'\|_0 \leq \rho$, and such that our empirical classification procedure misclassifies or abstains on $\vx'$. 
    \item The \textit{median adversarial attack magnitude} on a dataset is the median value of the \textit{empirical adversarial attack magnitude} across the dataset.
\end{itemize}
Unless otherwise stated, the uncertainty $\alpha$ is 0.05, and 10,000 randomly-ablated samples are used to make each prediction. The empirical estimation procedure we use to generate certificates, from \cite{cohen2019certified}, requires two sampling steps: the first to identify the majority class $i$, and the second to bound $\underline{p_i(\vx)}$. We use 1,000 and 10,000 samples, respectively, for these two steps.
\subsection{Results on MNIST}
We first tested our robust classification scheme on MNIST, using a simple CNN model as the base classifier (see appendix for architectural details.) Results are presented in Table \ref{mnistcert}. We varied the number of retained pixels $k$ in each sample: note that for small $k$, certified robustness and accuracy both increase as $k$ increases. However, after a certain threshold, here achieved at $k=45$, certified robustness starts to decrease with $k$, while classification accuracy continues to increase. This can be understood by considering Figure \ref{fig:bound-plot}: For larger $k$, the bounding constant $\Delta$ grows considerably faster with the $L_0$ distance $\rho$. In other words, a larger fraction of ablated samples must be classified correctly to achieve the same certified robustness. For small $k$, the fraction of ablated samples classified correctly increases sufficiently quickly with $k$ to counteract this effect; however, after a certain point, it is no longer beneficial to increase $k$ because a large majority of samples are already classified correctly by the base classifier (For example, see Figure \ref{fig:intro-fig}).\\
\begin{table} [ht]
\centering
\begin{tabular}{c|c|c}
Retained&Classification accuracy&Median certified\\
pixels $k$&(Percent abstained)&robustness\\
\hline
5&32.32\% (5.65\%)& N/A\\
10&74.90\% (5.08\%)&0\\
15&86.09\% (2.82\%)&0\\
20&90.29\% (1.81\%)&3\\
25&93.05\% (1.02\%)&5\\
30&94.68\% (0.77\%)&7\\
35&95.40\% (0.66\%)&7\\
40&96.27\% (0.52\%)&8\\
\textbf{45}&\textbf{96.72\% (0.45\%)}&\textbf{8}\\
50&97.16\% (0.32\%)&7\\
55&97.41\% (0.34\%)&7\\
60&97.78\% (0.18\%)&7\\
65&98.05\% (0.15\%)&6\\
70&98.18\% (0.20\%)&6\\
75&98.28\% (0.20\%)&6\\
80&98.37\% (0.12\%)&5\\
85&98.57\% (0.12\%)&5\\
90&98.58\% (0.16\%)&5\\
95&98.73\% (0.11\%)&5\\
100&98.75\% (0.16\%)&4\\
\end{tabular}
\caption{Robustness certificates on MNIST, using different numbers of retained pixels ($k$). The maximum median certified robustness on the MNIST test set is achieved when using $k=40$ or $k=45$ retained pixels: because $k=45$ gives better classification accuracy, we use this model (highlighted in bold) when evaluating against adversarial attacks. }
\label{mnistcert}
\end{table}
We also tested the empirical robustness of our classifier to an $L_0$ adversarial attack. Specifically, we chose to use the black-box \textit{Pointwise attack} proposed by \cite{schott2019towards}. We choose a black-box attack because comparisons to other robust classifiers using gradient-based attacks (such as the $L_0$ attack proposed by \cite{carlini2017towards}) may be somewhat asymmetric since our smoothed classifier is non-differentiable (because the base classifier's output is discretized.) While \cite{salman2019provably} does propose a gradient-based scheme for attacking $L_2$-smoothed classifiers which are similarly non-differentiable, adapting such a scheme would be a non-trivial departure from the existing $L_0$ Carlini-Wagner attack, precluding a direct comparison to other robust classifiers. By contrast, a practical reason we choose the Pointwise Attack is that the reference implementation of the attack is available as part of the Foolbox package \cite{rauber2017foolbox}, meaning that we can directly compare our results to that of \cite{schott2019towards}, without any concerns about implementation details. We note that \cite{schott2019towards} reports a median adversarial attack magnitude of 9 pixels for an unprotected CNN model on MNIST, which is comparable to the \textit{mean} adversarial attack magnitude of 8.5 reported for the $L_0$ Carlini-Wagner attack. This suggests that the attack is comparably effective. Results are presented in Table \ref{mnistemperical}. Note that our model appears to be significantly more robust to $L_0$ attack than any of the models tested by \cite{schott2019towards}, at a slight cost of classification accuracy (We would anticipate this trade-off, see \cite{tsipras2018robustness}.) Also note that while there is a gap between the median certified lower bound for the magnitude of any attack, 8 pixels, and the empirical upper bound given by an extant attack, 31 pixels, these quantities are at least in the same order of magnitude, indicating that our certificate is a non-trivial guarantee. See Figure 3 for examples of adversarial attacks on our classifier.
\begin{table} [ht]
\centering
\begin{tabular}{c|c|c}
Model&Class.&Median adv.\\
&acc.&attack mag.\\
\hline
CNN&99.1\% &9.0\\
Binarized CNN&98.5\% &11.0\\
Nearest Neighbor&96.9\% &10.0\\
$L_\infty$-Robust \cite{madry2017towards}&98.8\% &4.0\\
\cite{schott2019towards}&99.0\%&16.5\\
Binarized \cite{schott2019towards}&99.0\% &22.0\\
\textbf{Our model ($k=45$)}&\textbf{96.7\%} &\textbf{31.0}\\
\end{tabular}
\caption{Median adversarial attack magnitude on MNIST using the Pointwise attack from \cite{schott2019towards}, taking the best attack on each image from 10 random restarts. Note that all values except for our model are taken directly from \cite{schott2019towards}. For every evaluation performed by the black-box attack, 10,000 ablated samples were used to calculate class scores of our model: this was to ensure stability of the evaluated scores. Additionally, causing our model to abstain from classifying was counted as a successful attack, even if the correct class score was still marginally highest. Because the black-box attack performs a large number of classifications, and each of these classifications required 10,000 evaluations of the base classifier, we used only a subset of the MNIST test set, consisting of 275 images. }
\label{mnistemperical}
\end{table}
\begin{figure}[ht!]
    \centering
    \includegraphics[width=0.35\textwidth]{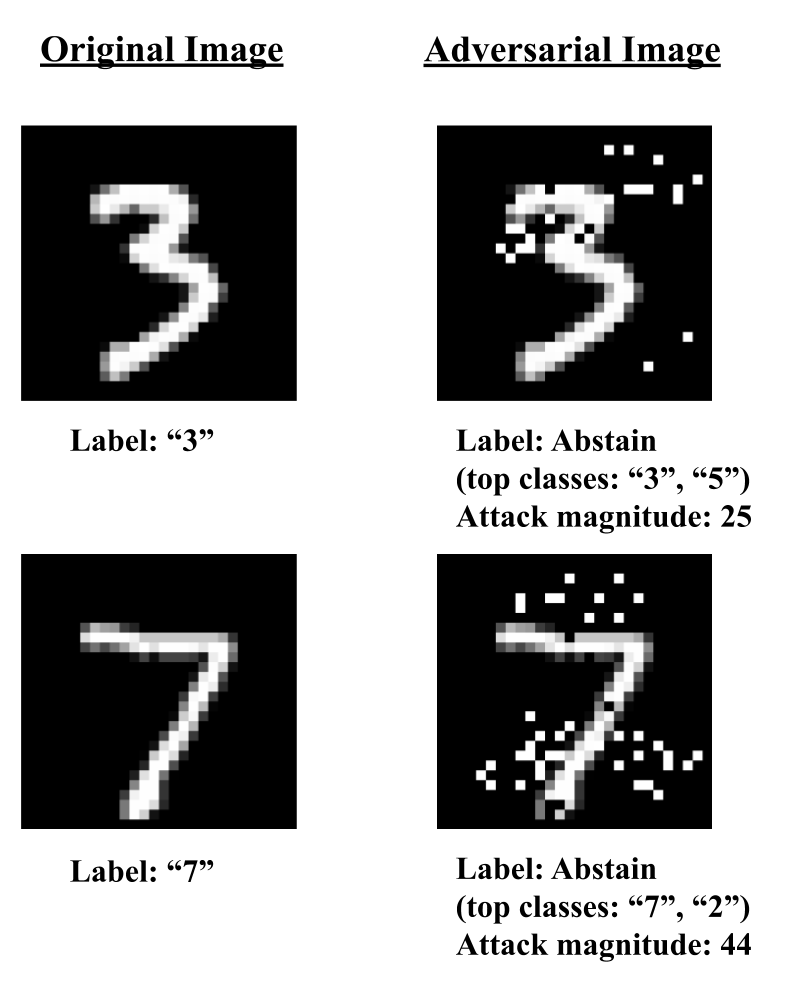}
    \caption{Adversarial examples to our classifier on MNIST. Note that because we consider the classifier abstaining to be a successful attack, these adversarial examples are in fact on the boundary between classes, rather than being entirely misclassified.}
    \label{fig:advexamp}
\end{figure}
\subsection{Results on CIFAR-10}
\begin{table}[ht]
\centering
\begin{tabular}{c|c|c}
Retained&Classification accuracy&Median certified\\
pixels $k$&(Percent abstained)&robustness\\
\hline
25&68.41\% (1.76\%)&6\\
50&74.21\% (1.19\%)&7\\
\textbf{75}&\textbf{78.25\% (0.93\%)}&\textbf{7}\\
100&80.91\% (0.86\%)&6\\
125&83.25\% (0.60\%)&5\\
150&85.22\% (0.53\%)&4\\
\end{tabular}
\caption{Robustness certificates on CIFAR-10, using different numbers of retained pixels ($k$), and using ResNet18 \cite{he2016deep} as the base classifier. Note that without smoothing, the base implementation of an unprotected ResNet18 classifier which we used \cite{torchcifar} has a classification accuracy of 93.02\% on CIFAR-10.}
\label{cifarcert}
\end{table}
\begin{table} [ht]
\centering
\begin{tabular}{c|c|c}
Retained&Base classifier&Base classifier\\
pixels $k$&training accuracy&test accuracy\\
\hline
25&83.16\%&57.72\%\\
50&96.63\%&68.29\%\\
75&99.33\%&74.08\%\\
100&99.76\%&77.88\%\\
125&99.91\%&80.48\%\\
150&99.95\%&83.16\%\\
\end{tabular}
\caption{Accuracy of the base classifier $f$ in CIFAR-10 experiments, on training versus test data, using ResNet18. Note that the base classifier significantly overfits to the training data. (Training accuracies are averaged over the final epoch of training.) }
\label{cifaroverfit}
\end{table}
\begin{table}[ht]
\centering
\begin{tabular}{c|c|c}
Retained&Base classifier&Base classifier\\
pixels $k$&training accuracy&test accuracy\\
\hline
25&83.89\%&57.58\%\\
50&96.91\%&69.45\%\\
75&99.09\%&75.22\%\\
100&99.66\%&79.54\%\\
125&99.78\%&81.83\%\\
150&99.92\%&84.43\%\\
\end{tabular}
\caption{Accuracy of the base classifier $f$ in CIFAR-10 experiments, on training versus test data, using ResNet50. Note that the base classifier significantly overfits to the training data: however, for $k > 25$, this higher-capacity model overfits less than ResNet18. }
\label{cifaroverfit50}
\end{table}
We implemented our technique on CIFAR-10 using ResNet18 (with the number of input channels increased to 6) as a base classifier; see Table \ref{cifarcert} for our robustness certificates as a function of $k$. The median certified robustness is somewhat smaller than for MNIST: however, this is in line with the performance of empirical attacks. For example, the $L_0$ attack proposed by \cite{carlini2017towards} achieves a mean adversarial attack magnitude of 8.5 pixels on MNIST and 5.9 pixels on CIFAR-10. This suggests that CIFAR-10 samples are more vulnerable to $L_0$ adversarial attacks compared to the MNIST ones. Intuitively, this is because CIFAR-10 images are both visually complex and low-resolution, so that each pixel carries a large amount of information regarding the classification label. Also note that the classification accuracy on unperturbed images is somewhat reduced. For example, in a model using $k=150$, the median certified robustness is $4$ pixels, and the classifier accuracy is $85.22\%$. The trade-off between accuracy and robustness is also more pronounced. However, it is not unusual for practical $L_0$ defenses to achieve accuracy below 90\% on CIFAR-10 \cite{meng2017magnet,xu2017feature}: our defense may therefore still prove to be usable.\\
One phenomenon which we encountered when applying our technique to CIFAR-10 was over-fitting of the base classifier (see Table \ref{cifaroverfit}), which was unexpected because during the training, the classifier is always exposed to new random ablations of the training data. However, the network was still able to memorize the training data, despite never being exposed to the complete images. While interpolation of even randomly labeled training data is a known phenomenon in deep learning \cite{zhang2016understanding}, we were surprised to see that over-fitting may happen on ablated images, where a particular ablation is likely never repeated in training. In order to better understand this, we use a model trained on a higher-capacity network architecture, ResNet50. The results for the base classifier are given in Table \ref{cifaroverfit50}. Surprisingly, increasing network capacity decreased the generalization gap slightly for $k \geq 50$ (Note that because the improvement to the base classifier is only marginal, and because ResNet50 is substantially more computationally intensive to use as a base classifier to classify 10,000 ablated samples per image, we opted to compute certificates using the ResNet18 model).
\subsection{Results on ImageNet}
We implemented our technique on ImageNet using ResNet50 (again with the number of input channels increased to 6) as a base classifier; see Table \ref{imagenetcert} for our robustness certificates as a function of $k$. For testing, we used a random subset of 400 images from the ILSVRC2012 validation set. Note that ImageNet classification is a 1,000-class problem: here we consider only top-1 accuracy. 
Because these top-1 accuracies are only moderately above 50 percent, the calculation of the median certified robustness is skewed by relatively large fraction of misclassified points: on the points which are correctly classified, the certificates can be considerably larger. For example, at $k=1000$, if we consider only the 61\% of images which are certified for the correct class, the median certificate is {\it 33 pixels}. Similarly, considering only images with certificates other than `N/A', the median certificates for $k=500$ and $k=2000$ are 63 pixels and 16 pixels, respectively.
\begin{table}[h!] 
\centering
\begin{tabular}{c|c|c}
Retained&Classification accuracy&Median certified\\
pixels $k$&(Percent abstained)&robustness\\
\hline
500&52.75\% (1.75\%)&0\\
\textbf{1000}&\textbf{61.00\% (0.00\%)}&\textbf{16}\\
2000&62.50\% (1.75\%)&11\\
\end{tabular}
\caption{Robustness certificates on ImageNet, using different numbers of retained pixels $k$, and using ResNet50 \cite{he2016deep} as the base classifier.  For ImageNet,  $d=224 \times 224$. Note that without smoothing, the base implementation of an unprotected ResNet50 classifier can be trained on ImageNet to a top-1 accuracy of 76.15\% \cite{paszke2017automatic}.}
\label{imagenetcert}
\end{table}
\section{Discussion}
\subsection{Comparison to \cite{lee2019tight}}
In a concurrent work, \cite{lee2019tight} also present a randomized-smoothing based robustness certification scheme for the $L_0$ metric. In this scheme, each pixel is retained with a fixed probability $\kappa$ and is otherwise assigned to a \textit{random} value from the remaining possible pixel values in $\gS$. Note that there is no $\nullsim$ in this scheme. As a consequence, the base classifier lacks explicit information about \textit{which} pixels are retained from the original image, and which have been randomized. The resulting scheme has considerably lower median certified robustness on the datasets tested in both works\footnote{\cite{lee2019tight} uses a similar scheme to ours to derive an empirical bound on $\underline{p_i(\vx)}$; however, that work uses 100 samples to select $i$ and 100,000 samples to bound it, and reports bounds with 99.9\% confidence ($\alpha = .001$). In order to provide a fair comparison, we repeated our certifications on MNIST and ImageNet (for optimized values of $k$) using these empirical certification parameters. This did not change the median robustness certificates. } (Table \ref{vslee}):

\begin{table}[h!] 
\centering
\begin{tabular}{c|c|c}
Dataset&Median certified&Median certified\\
&robustness (pixels)&robustness (pixels)\\
&\cite{lee2019tight}&(our model)\\
\hline
MNIST&4&\textbf{8}\\
ImageNet&1&\textbf{16}\\
\end{tabular}
\caption{Comparison of robustness certificates in \cite{lee2019tight} and in this work, using the optimal choices of hyperparameters tested in each work. Numbers for \cite{lee2019tight} are derived from those reported in that work. Note that for ImageNet, \cite{lee2019tight} considers each color channel as a separate pixel: therefore the median image is robust to distortion in only \textit{one channel} of one pixel. By contrast, our model is robust to distortions in \textit{all channels} in 16 pixels (or, in the limiting case, one channel in 16 pixels). }
\label{vslee}
\end{table}
To illustrate quantitatively how our robust classifier obtains more information from each ablated sample than is available in the \textit{randomly noised} samples in \cite{lee2019tight}, let us consider images of ImageNet scale. Because \cite{lee2019tight} considers each color channel as a separate pixel when computing certificates, we will use $\gS = \{0,...,255\}$, and $d = 3*224*224$. Using \cite{lee2019tight}'s certificate scheme, in order to certify for one pixel of robustness with $\kappa=0.1$ probability of pixel retention, we would need to accurately classify noised images with probability $p_i(\vx) = .596$. Meanwhile, using our ablation scheme, in order to certify one pixel of robustness by correctly classifying same fraction ($p_i(\vx) = .596$) of ablated images, we can retain at most $k=14521$ pixels. This is $9.6\%$ of pixels, slightly fewer than the expected number retained in \cite{lee2019tight}'s scheme.

However, we will now calculate the \textit{mutual information} between each ablated/noised image and the original image for each scheme: this is the expected number of bits of information about the original image which are obtained from observing the ablated/noised image. For illustrative purposes, we will make the simplifying assumption that the dataset overall is uniformly distributed (while this is obviously not true for image classification, it is a reasonable assumption in other classification tasks.) In our scheme, we have simply 
\begin{equation}
    I_{\text{ablate}} = \log_2|\gS|*k = 8*k = 116168\text{ bits.}
\end{equation}
Each of the $k$ retained pixels provides $8$ bits of information. However, in the noising scheme from \cite{lee2019tight}, we instead have:
\begin{equation} \label{eq:leemutinfo}
\begin{split}
       &I_{\text{Lee et al.}}\\
      &= d\left(\log_2|\gS| + \kappa\log_2\kappa + (1-\kappa)\log_2\frac{1-\kappa}{|\gS|-1} \right)\\
       &\approx 50590.4\text{ bits.} 
\end{split}
\end{equation}
Therefore, despite using slightly fewer pixels from the original image, over twice the amount of information about the original image is available in our scheme when making each ablated classification. (A derivation of Equation \ref{eq:leemutinfo} is provided in the appendix.)
\subsection{Alternative encodings of $\gS_{\nullsim}$}
\begin{figure}[h]
    \centering
    \includegraphics[width=0.45\textwidth]{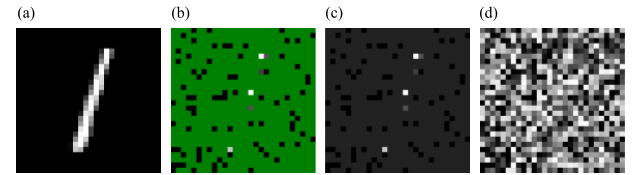}
    \caption{(a) An image from MNIST. (b) The image with $k=85$ pixels ablated, with a unique $\nullsim$ encoding. (c) The same image with $\nullsim$ encoded as the mean pixel value (dark grey). Note that both black and white pixels are still distinguishable. (d) If we replace ablated pixels with random noise, the image is no longer easily distinguishable. }
    \label{fig:encodings}
\end{figure}

The multichannel encoding of $\gS_{\nullsim}$ described above, while theoretically well-motivated, is not the only possible encoding scheme. In fact, for MNIST and CIFAR-10, we tested a somewhat simpler encoding for the $\nullsim$ symbol: we simply used the mean pixel value on the training set, similarly to the practical defense proposed by  \cite{hosseini2019dropping}. We tested using the optimal values of $k$ from the Results section above ($k=45$ for MNIST and $k=75$ for CIFAR-10). This resulted in only marginally decreased accuracy and certificate sizes (Table \ref{snulltable}):
\begin{table}[hb] 
\centering
\begin{tabular}{c|c|c}
$\gS_{\nullsim}$&Classification acc.&Median certified\\
encoding&(Pct. abstained)&robustness\\
\hline
\textbf{MNIST}&&\\
\hline
Multichannel&\textbf{96.72\% (0.45\%)}&\textbf{8}\\
Mean &96.27\% (0.43\%)&7\\
\hline
\textbf{CIFAR-10}&&\\
\hline
Multichannel&\textbf{78.25\% (0.93\%)}&\textbf{7}\\
Mean&77.71\% (1.05\%)&\textbf{7}\\
\end{tabular}
\caption{Accuracy and robustness using different encoding schemes for $\gS_{\nullsim}$. }
\label{snulltable}
\end{table}\\
To understand this, note that the \textit{mean} pixel value (grey in both datasets) is not necessarily a \textit{common} value: it is still possible to distinguish which pixels are ablated (Figure \ref{fig:encodings}).
\section{Conclusion}
 In this paper, we introduced a novel smoothing-based certifiably robust classification method against sparse adversarial attacks, in which the adversary can perturb a certain number features in input samples. Our method, which is modeled after randomised smoothing methods for certifiably robust classification for $L_1$ and $L_2$ attack models, was shown to produce non-trivial robustness certificates on MNIST, CIFAR-10, and ImageNet, and to be an effective empirical defense against $L_0$ attacks on MNIST.
\section{Acknowledgements}
This work was supported in part by NSF award CDS\&E:1854532 and award HR00111990077.
\fontsize{9.5pt}{10.5pt} \selectfont
\bibliographystyle{aaai}
\bibliography{references}
\newpage
\appendix 
\nocopyright 
\section{Architecture and Training Parameters for MNIST}
See Tables \ref{tab:MNISTlayers} and \ref{tab:MNISTtrain}.
\begin{table}[h!]
    \centering
    \begin{tabular}{|c|c|}
    \hline
        Layer & Output Shape\\
        \hline
        (Input) & $2 \times 28 \times 28$ \\
        2D Convolution + ReLU & $64 \times 14 \times 14 $\\
        2D Convolution + ReLU & $128 \times 7 \times 7$ \\
        Flatten & 6272 \\
        Fully Connected + ReLU & 500 \\
        Fully Connected + ReLU & 100 \\
        Fully Connected + SoftMax & 10 \\
        \hline
    \end{tabular}
    \caption{Model Architecture of the Base Classifier for MNIST Experiments. 2D Convolution layers both have a kernel size of 4-by-4 pixels, stride of 2 pixels, and padding of 1 pixel.}
    \label{tab:MNISTlayers}
\end{table}
\begin{table}[h!]
    \centering
    \begin{tabular}{|c|c|}
        \hline
        Training Epochs & 400\\
        \hline
        Batch Size & 128\\
        \hline
        Optimizer & Stochastic Gradient \\
        &Descent with Momentum\\
        \hline
        Learning Rate & .01 (Epochs 1-200) \\
        & .001 (Epochs 201-400)\\
        \hline
        Momentum & 0.9 \\
        \hline
        $L_2$ Weight Penalty & 0 \\
        \hline
    \end{tabular}
    \caption{Training Parameters for MNIST Experiments}
    \label{tab:MNISTtrain}
\end{table}
\section{Training Parameters for CIFAR-10}
As discussed in the main text, we used a standard ResNet18 architecture for our base classifier: the only modification made was to increase the number of input channels from 3 to 6. See Table \ref{tab:CIFARtrain} for training parameters.
\begin{table}[h!]
    \centering
    \begin{tabular}{|c|c|}
        \hline
        Training Epochs & 400\\
        \hline
        Batch Size & 128\\
        \hline
        Training Set & Random Cropping (Padding:4)\\
        Preprocessing & and Random Horizontal Flip\\
        \hline
        Optimizer & Stochastic Gradient \\
        &Descent with Momentum\\
        \hline
        Learning Rate & .01 (Epochs 1-200) \\
        & .001 (Epochs 201-400)\\
        \hline
        Momentum & 0.9 \\
        \hline
        $L_2$ Weight Penalty & 0.0005 \\
        \hline
    \end{tabular}
    \caption{Training Parameters for CIFAR-10 Experiments}
    \label{tab:CIFARtrain}
\end{table}
\section{Training Parameters for ImageNet}
As with CIFAR-10, we used a standard ResNet50 architecture for our base classifier: the only modification made was to increase the number of input channels from 3 to 6. See Table \ref{tab:Imgnttrain} for training parameters.
\begin{table}[t!]
    \centering
    \begin{tabular}{|c|c|}
        \hline
        Training Epochs & 36\\
        \hline
        Batch Size & 256\\
        \hline
        Training Set & Random Resizing and Cropping,\\
        Preprocessing & Random Horizontal Flip\\
        \hline
        Optimizer & Stochastic Gradient \\
        &Descent with Momentum\\
        \hline
        Learning Rate & .1 (21 Epochs) \\
        & .01 (10 Epochs) \\
        & .001 (5 Epochs) \\
        \hline
        Momentum & 0.9 \\
        \hline
        $L_2$ Weight Penalty & 0.0001 \\
        \hline
    \end{tabular}
    \caption{Training Parameters for ImageNet Experiments}
    \label{tab:Imgnttrain}
\end{table}
\section{Mutual information derivation for Lee et al. 2019}
Here we present a derivation of the expression given in Equation 21 in the main text. Let $\mathbf{X}$ be a random variable representing the original image: in this derivation, we assume that $\mathbf{X}$ is distributed uniformly in $\gS^d$. Let  $\mathbf{Y}$ be a random variable representing the image, after replacing each pixel with a random, different value with probability $(1-\kappa)$. By the definition of mutual information, we have:
\begin{equation}
I(\mathbf{X},\mathbf{Y}) = H(\mathbf{X}) - H(\mathbf{X}|\mathbf{Y})
\end{equation}
Note that, with $\mathbf{X}$ distributed uniformly, it consists of $d$ i.i.d. instances of a random variable $X_\circ$, itself uniformly distributed in $\gS$. Similarly, each component of $\mathbf{Y}$ is an instance of a random variable defined by:
\begin{equation}
   Y_\circ  =
\left\{
	\begin{array}{ll}
		X_\circ  & \mbox{with probability } \kappa \\
		\mbox{Uniform on }\gS-\{X_\circ\} &  \mbox{with probability } 1-\kappa
	\end{array}
\right.
\end{equation}
We can then factorize the expression for mutual information, using the fact that each instance of $(X_\circ,Y_\circ)$ is independent:
\begin{equation}
I_\text{Lee et al.} = I(\mathbf{X},\mathbf{Y}) = d(H(X_\circ) - H(X_\circ|Y_\circ))
\end{equation}
By the definitions of entropy and mutual entropy, we have:
\begin{equation}
\begin{split}
& I_\text{Lee et al.}= -d\biggl(\sum_{s\in\gS}  \Pr(X_\circ = s) \log_2 \Pr(X_\circ = s) \\
& - \sum_{(s,s')}  \Pr(X_\circ = s,Y_\circ = s') \log_2 \frac{\Pr(X_\circ = s,Y_\circ = s') }{\Pr(Y_\circ = s')}\biggr)  
\end{split}
\end{equation}
Note that, by symmetry, $Y_\circ$ is itself uniformly distributed on $\gS$. Then we have:
\begin{equation}
\begin{split}
& I_\text{Lee et al.}= -d\biggl(\sum_{s\in\gS} |\gS|^{-1} \log_2 |\gS|^{-1} \\
& - \sum_{(s,s')}  \Pr(X_\circ = s,Y_\circ = s') \log_2 \frac{\Pr(X_\circ = s,Y_\circ = s') }{|\gS|^{-1}}\biggr)  
\end{split}
\end{equation}
Splitting $(s,s')$ into cases for $(s=s')$ and $(s\neq  s')$:
\begin{equation} \label{eq:leeinfoint}
\begin{split}
& I_\text{Lee et al.}= -d\biggl(\sum_{s} |\gS|^{-1} \log_2 |\gS|^{-1} \\
& - \sum_{s}  \Pr(X_\circ = Y_\circ = s) \log_2 \frac{\Pr(X_\circ = Y_\circ = s) }{|\gS|^{-1}}\\
&- \sum_{s \neq s'}  \Pr(X_\circ =s, Y_\circ = s') \log_2 \frac{\Pr(X_\circ =s,  Y_\circ = s') }{|\gS|^{-1}}\biggr)  
\end{split}
\end{equation}
Note that $\Pr(X_\circ = Y_\circ = s) $ = $|\gS|^{-1}\kappa$, because $X_\circ = s$ with probability $|\gS|^{-1}$, and then $Y_\circ$ is assigned to $X_\circ$ with probability $\kappa$. Also, for $s \neq s'$, we have
\begin{equation}
\Pr(X_\circ =s, Y_\circ = s') = |\gS|^{-1}(1-\kappa)(|\gS|-1)^{-1},
\end{equation}
because $X_\circ = s$ with probability $|\gS|^{-1}$, $Y_\circ$ is not equal to $X_\circ$  with probability  $(1-\kappa)$, and then $Y_\circ$ assumes each value in $\gS-\{X_\circ\}$ with uniform probability. Plugging these expressions into Equation \ref{eq:leeinfoint} gives:
\begin{equation} 
\begin{split}
& I_\text{Lee et al.}= -d\biggl(\sum_{s} \frac{ \log_2 |\gS|^{-1}}{|\gS|}  - \sum_{s} \frac{\kappa}{|\gS|}  \log_2 \kappa\\
&- \sum_{s \neq s'} \frac{ (1-\kappa)}{(|\gS|-1)|\gS|}\log_2 \left[(1-\kappa)(|\gS|-1)^{-1}\right]\biggr)  
\end{split}
\end{equation}
Now all summands are constants: we note that summing over all $s\in \gS$ is now equivalent to multiplying by $|\gS|$ and summing over $(s,s') \in \gS^2$ with $s \neq s'$ is equivalent to multiplying by $|\gS|(|\gS|-1)$:
\begin{equation} 
\begin{split}
& I_\text{Lee et al.}= -d\bigl( \log_2 |\gS|^{-1}  - \kappa \log_2 \kappa\\
&-   (1-\kappa)\log_2 \left[(1-\kappa)(|\gS|-1)^{-1}\right]\bigr)  
\end{split}
\end{equation}
This simplifies to the expression given in the text.
\section{Additional Adversarial Examples}
\begin{figure*}[ht!]
    \centering
    \includegraphics[width=.9\textwidth]{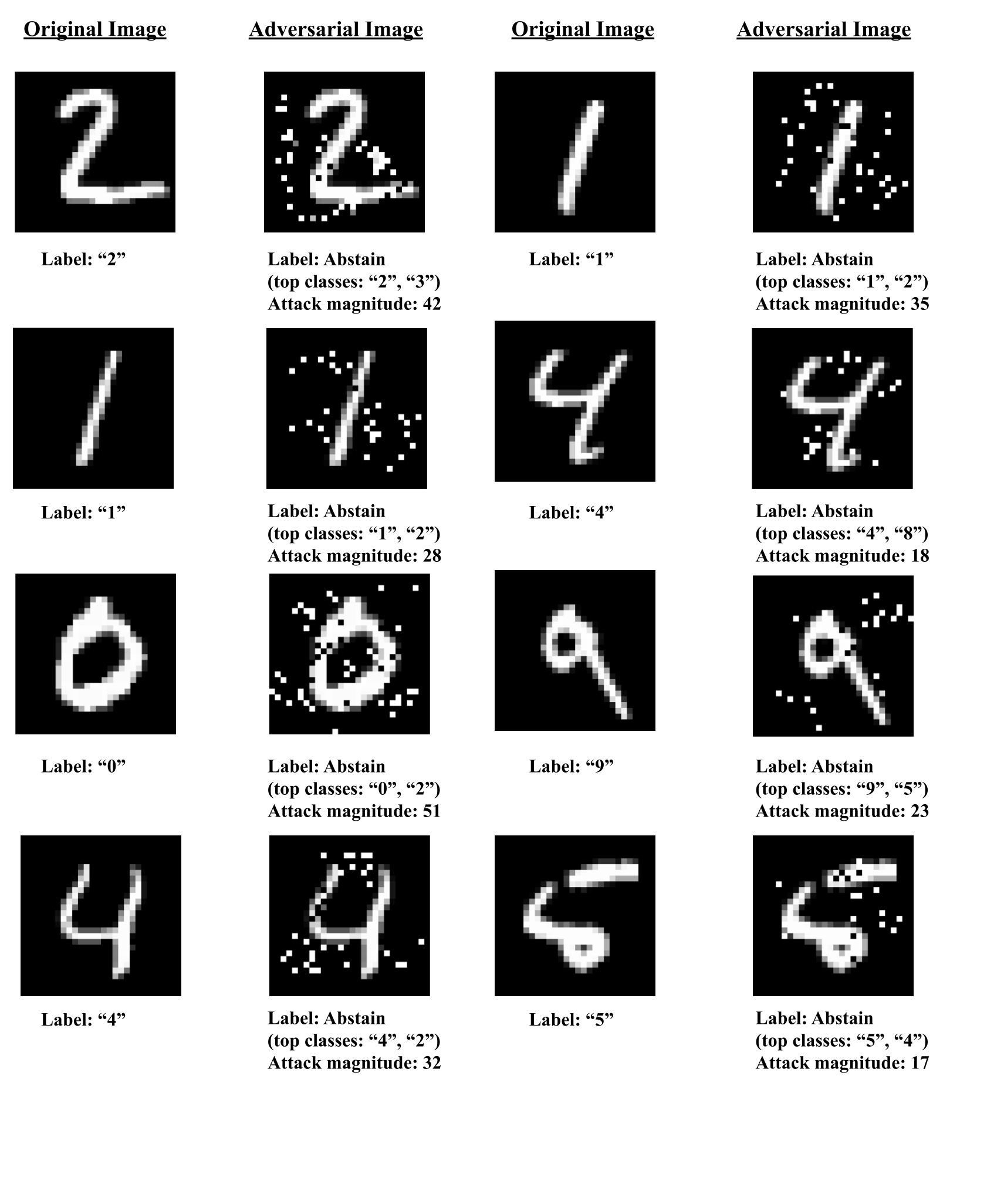}
    \caption{Additional adversarial examples generated on MNIST by the Pointwise attack on our robust classifier, with $k=45$. }
    \label{fig:suppfig}
\end{figure*}

See Figure \ref{fig:suppfig}.
\end{document}